\documentclass[10pt,twocolumn,letterpaper]{article}

\usepackage{iccv}
\usepackage{times}
\usepackage{epsfig}
\usepackage{graphicx}
\usepackage{amsmath}
\usepackage{amssymb}

\usepackage[accsupp]{axessibility} 
\usepackage{bm}
\usepackage{amsthm}
\usepackage{mathrsfs}
\usepackage{color}
\usepackage{multirow}
\usepackage{makecell}
\usepackage{subfigure}
\makeatletter
\@namedef{ver@everyshi.sty}{}
\makeatother
\usepackage{tikz}

\usetikzlibrary{spy}

\usepackage[pagebackref=true,breaklinks=true,letterpaper=true,colorlinks,bookmarks=false]{hyperref}

\iccvfinalcopy 

\def\iccvPaperID{7727} 

\ificcvfinal\fi

\newcommand{\myname}[0]{{RED}}

\begin{document}

\title{Motion Deblurring with Real Events}
\author{Fang Xu$^{1}$, Lei Yu$^{1}$\footnotemark[1], Bishan Wang$^{1}$, Wen Yang$^{1}$\footnotemark[1], Gui-Song Xia$^{2}$,
Xu Jia$^{3}$\footnotemark[1], Zhendong Qiao$^{4}$, Jianzhuang Liu$^{4}$\\
$^1$School of Electronic Information, Wuhan University\\
$^2$School of Computer Science, Wuhan University\\
$^3$School of Artificial Intelligence, Dalian University of Technology\\
$^4$Noah's Ark Lab, Huawei Technologies\\
{{\small \{xufang, ly.wd, bswang, yangwen, guisong.xia\}@whu.edu.cn, xjia@dlut.edu.cn, \{qiaozhendong, liu.jianzhuang\}@huawei.com}}
}
\maketitle
\ificcvfinal\thispagestyle{empty}\fi

\begin{abstract}
\vspace{-0.3cm}
In this paper, we propose an end-to-end learning framework for event-based motion deblurring in a self-supervised manner, where real-world events are exploited to alleviate the performance degradation caused by data inconsistency. To achieve this end, optical flows are predicted from events, with which the blurry consistency and photometric consistency are exploited to enable self-supervision on the deblurring network with real-world data. Furthermore, a piece-wise linear motion model is proposed to take into account motion non-linearities and thus leads to an accurate model for the physical formation of motion blurs in the real-world scenario. Extensive evaluation on both synthetic and real motion blur datasets demonstrates that the proposed algorithm bridges the gap between simulated and real-world motion blurs and shows remarkable performance for event-based motion deblurring in real-world scenarios.
\end{abstract}
\vspace{-0.5cm}

\section{Introduction}
\footnotetext[1]{Corresponding author}
\footnotetext{The research was partially supported by the National Natural Science Foundation of China, No. 61871297 and the Fundamental Research Funds for the Central University of China, WHU No. 2042020kf0019 and DUT No. 82232026. And the numerical calculations have been done on the supercomputing system in the Supercomputing Center of Wuhan University.
}

Due to motion ambiguities as well as the erasure of intensity textures \cite{jin2018learning}, the task of motion deblurring is severely ill-posed \cite{jin2019learning,purohit2019bringing}. With the help of an event camera which can ``continuously'' emit events asynchronously with extremely low latency (in the order of $\mu$s) \cite{lichtsteiner128Times1282008,gallego2020event}, inherently embedding motions and textures~\cite{benosman2013event}, the task of motion deblurring \cite{pan2019bringing,pan2020high,scheerlinck2018continuous}  can be essentially alleviated. Many event-based motion deblurring methods have been proposed by learning from synthesized dataset composed of simulated events and blurry images as well as sequences of sharp clear ground-truth images \cite{jiang2020learning,wang2020event}. However, the inconsistency between synthetic and real data degrades the performance of inference on real-world event cameras \cite{stoffregen2020reducing}.
\begin{figure}[!t]
    \centering
    \includegraphics[width=.48\textwidth]{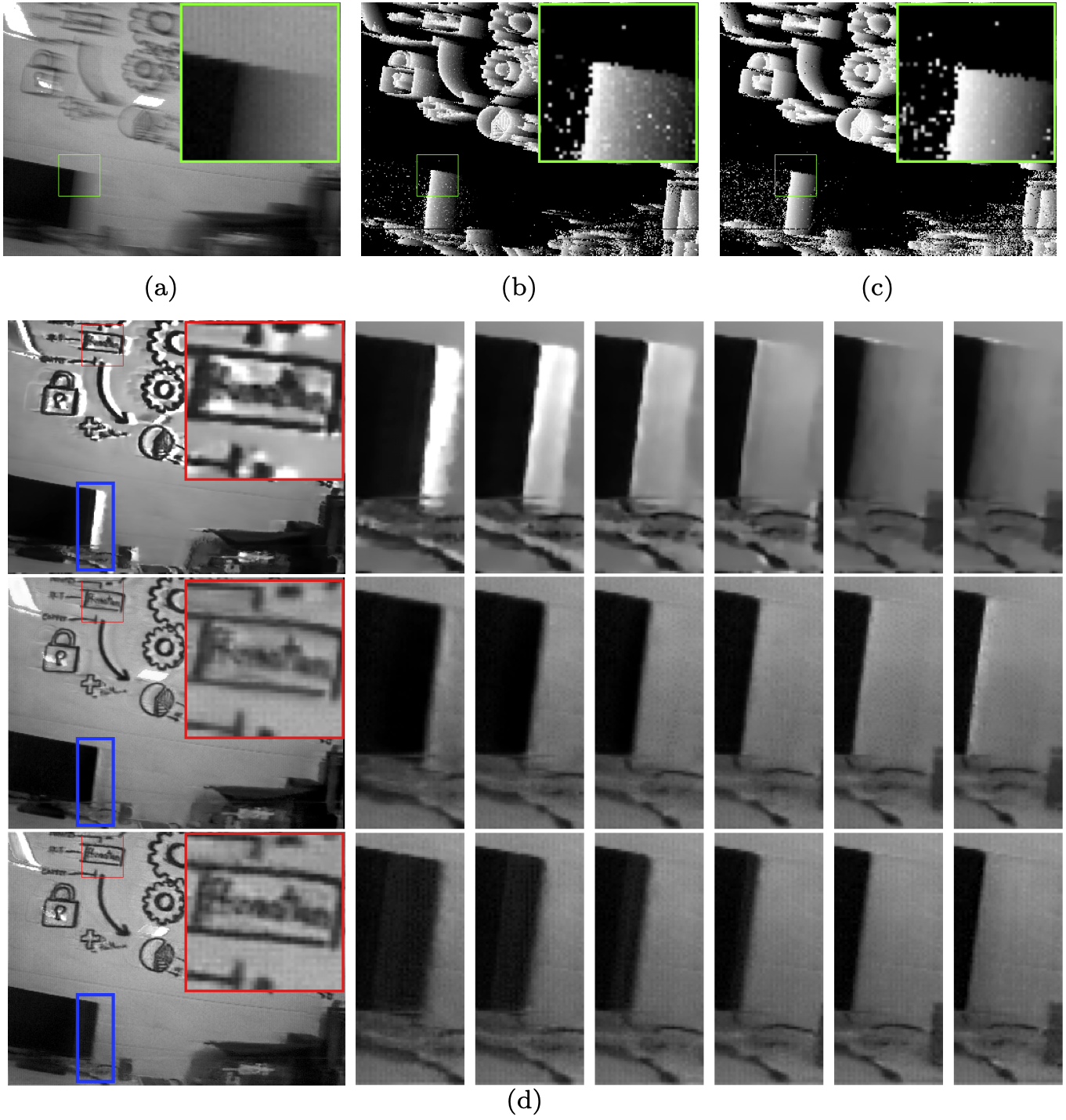}
    \caption{Illustrative example of inconsistency between synthesized and real-world motion blurs with respect to the event time-surface \cite{lagorce2016hots}: (a) a real-world motion blurred image; (b) the time-surface of real-world events corresponding to (a); (c) the time-surface of events collected with the same trajectory as (a) but at a slow motion speed; (d) deblurred results respectively by eSL-Net \cite{wang2020event} (top row), LEDVDI \cite{lin2020learning} (middle row) and our proposed RED-Net (bottom row) that trained with real-world events and real-world motion blurs, sequence predictions of the blue area are shown on the right. Our proposed RED-Net generates less {\it halo artifacts} and achieves  the best visualization performance.}
    \label{first}
    \vspace{-0.5cm}
\end{figure}

The physical intrinsic noise of event cameras raises the difficulty of simulating labeled events that highly match the real event data \cite{rebecq2018esim}. Even though the event simulator to some extent reduces the gap by considering the pixel-to-pixel variation in the event threshold \cite{rebecq2018esim}, additional noise effects such as background activity noise and false negatives \cite{baldwin2020event} still exist, leading to tremendous discrepancy between the virtual events synthesized from event simulators \cite{rebecq2018esim} and the real events emitted by event cameras. An alternative approach is to build a labeled dataset composed of real-world events accompanying with synthesized blurry images, and then train networks on it \cite{jiang2020learning}. Unfortunately, obtaining such pairs is not always easy, which needs to be captured with a slow motion speed 
as well as under good lighting conditions to avoid motion blur. Subsequent blurry image synthesis and alignment on temporal domain is also tedious but indispensable. Furthermore, inconsistency still exists between the events associated with synthesized and real-world motion blur in that limited read-out bandwidth leads to more event timing variations, as shown in Fig.~\ref{first}.  

In this paper, we propose a novel framework of learning the event-based motion deblurring network in a self-supervised manner, where real-world events and real-world motion blurred images are exploited to alleviate the performance degradation caused by data inconsistency and bridge the gap between simulations and real-world scenario. The proposed framework consists of two neural networks, an event-based motion deblurring network (Deblur-Net) and an event-based optical flow estimation network (OF-Net). The Deblur-Net is fed with both events and a single motion blurred image, and outputs a sequence of sharp clear images, while the OF-Net receives events and provides motions between the reconstructed sharp clear images. We relate motions and sharp clear images according to the photometric constancy \cite{zhu2018ev}. The real-world motion blurs with non-linearities are considered by a piece-wise linear motion (PLM) model which improves the accuracy of optical flow and thus provides a more precise blurring model between the reconstructed intensity images from Deblur-Net and the blurry input. The overall network is jointly trained end-to-end over partially labeled dataset composed of synthetic data using ESIM \cite{rebecq2018esim} with ground-truth sharp clear images and real-world data only containing real-world events and real-world motion blurred images, which can be captured simultaneously by the DAVIS \cite{brandli2014b} or with a dual camera set connecting the event camera and the RGB camera with a beam splitter \cite{wang2020joint}. 

The main contributions of our work are in three folds:
\begin{itemize} 
	\item We propose a framework of learning the event-based motion deblurring network with real-world events, which remarkably improves the performance of motion deblurring in the real-world scenario. 
	\item We propose a piece-wise linear motion model to consider the motion nonlinearity, based on which the OF-Net is ameliorated for real events and outputs accurate and dense motion flows.
	\item Extensive experiments show that the proposed method can yield high quality sharp frames and achieve state-of-the-art results on the real blurry event dataset. 
\end{itemize}                                     

\section{Related Work}
\noindent{\bf Motion Deblurring.}
Even though motion blur degenerates photographs by averaging over the exposure time intervals, it inherently embeds both motions and textures of the moving objects and thus enables the possibility of motion deblurring for visualization of the dynamic scene behind blurred photographs \cite{jin2018learning,jin2019learning,purohit2019bringing}. 
Early attempts address this problem by assuming spatially uniform motion where blurs can be modelled as the convolution of a blur kernel with a sharp image, leading to the kernel-based motion deblurring approaches by means of regularized deconvolution \cite{krishnan2011blind,sun2013edge}. For motion blurs caused by complex motion behaviors, pixel/patch-wise motion flows are often required to flexibly depict non-uniform motions. To achieve this end, the optical flow is often estimated from a single blurry image~\cite{gong2017motion} or consecutive frames~\cite{zhang2015intra} and provides photometric constancy \cite{hyun2015generalized} between the recovered latent sharp images, leading to flow-based motion deblurring approaches. Motion blurs bring the ambiguity of temporal ordering as well as the erasure of spatial textures \cite{jin2018learning}, thus several priors have been successfully employed for blur kernels, \eg, sparsity \cite{xu2013unnatural}, Gaussian scale mixture \cite{fergus2006removing}, and dark channel \cite{pan2016blind}, and for optical flows, \eg, linear motion \cite{liu2020self}, and order invariance \cite{jin2018learning}, which however may generate artifacts and degrade deblurring performance if the above assumptions are not fulfilled \cite{zhou2019spatio}. 

An alternative approach is to learn a neural network for motion deblurring directly from the data, which often achieves prominent performance \cite{jin2018learning,jin2019learning,purohit2019bringing,gong2017motion,nimisha2017blur}. Jin \etal pioneer to recover a sequence of sharp frames from a single motion-blurred image~\cite{jin2018learning}, where they sequentially train multiple neural networks with a temporal invariant loss. Purohit \etal utilize a single recurrent neural network to generate the entire sequence~\cite{purohit2019bringing} which, however, may suffer from ambiguities in temporal ordering. Take this into account, Jin \etal tackle the problem of the temporal ambiguity by feeding multiple motion blurred images~\cite{jin2019learning}, while Rengarajan \etal additionally use two consecutive images over a short exposure time \cite{rengarajan2020photosequencing}. On the other hand, Chen \etal propose a Reblur2Deblur network \cite{chen2018reblur2deblur} for motion deblurring via self-supervised learning to leverage the physical model and the data prior, where multiple blurry images and a linear motion assumption between them are essentially required. 

\noindent{\bf Event-Based Motion Deblurring.}
Event camera measures per-pixel brightness change and outputs an event once the change exceeds a threshold~\cite{gallego2020event}. The triggered events can be ``continuously'' emitted asynchronously with extremely low latency and thus provide missing information during exposure intervals if the motion blur happens \cite{pan2019bringing}. The temporal ambiguity and texture erasure can be easily tackled by introducing events into the deblurring algorithms \cite{pan2019bringing,lin2020learning}. Event-based motion deblurring methods can be categorized into two groups, \ie, {\it model driven} and {\it data driven} algorithms. Model-based algorithms relate events, blurry images and the corresponding latent sharp clear images according to the physical event generation principle \cite{pan2019bringing,pan2020single,scheerlinck2018continuous}. However, due to the imperfections of physical implementation including intrinsic noise and limited read-out bandwidth \cite{inivaiton2020,gallego2020event}, real events are essentially with noise both in temporal and spatial domains \cite{lichtsteiner128Times1282008} which inevitably degrades performance \cite{scheerlinck2018continuous}. 

Data driven algorithms relax the above limitations by utilizing  neural networks and directly learn the relation from a blurry image to a sequence of sharp clear images with the aid of events \cite{lin2020learning,wang2020event,jiang2020learning}. For training purpose, a synthesized dataset composed of labeled events and blurry images is commonly simulated from sharp clear video sequences \cite{jiang2020learning,wang2020event} which, however, may have inconsistency to the real events due to event noise in the spatio-temporal domain. Even though data variations have been considered by manually adding noise \cite{wang2020event,stoffregen2020reducing}, the generalizability remains limited for real event cameras. 

Jiang \etal \cite{jiang2020learning} build the Blur-DVS dataset based on real events triggered at a slow motion speed (to obtain a sequence of sharp clear intensity images as the ground truth) and train their network on it, where the sim-to-real gap is reduced to some extent \cite{lin2020learning}. However, building such a dataset is strenuous and requires strict conditions to obtain blur-free ground-truths, \eg, relative slow motion between camera and scenario and good lighting environment. Besides, due to the limited read-out bandwidth, the inconsistency between events emitted at a slow motion speed and their counterparts at a fast motion speed also exists. 

Learning with real-world data is more adaptive to the real-world scenario than synthetic data, and moreover, real-world events and motion blurs can be easily captured at extremely low cost without sophisticated procedures, which motivates us to propose a new framework to learn event-based motion deblurring by leveraging the {\it real-world events} and {\it real-world motion blurred images}.  

\section{Problem Statement}\label{sec:problem}
{\it Event-based motion deblur} (ED) aims at reconstructing a sequence of sharp and clear latent images $\{{\bf I}_{t}\}_{t\in T}$ from a single motion blurred image $\bf B$ captured with exposure time $T$ and corresponding events ${\bf E}_T \triangleq \{({\bf x}_i,p_i,t_i)\}_{t_i\in T}$ triggered in $T$ where $t_i$ and ${\bf x}_i$ respectively denote the timestamp and the pixel location of the $i$-th event, and $p_i\in \{+1,-1\}$ is the polarity. Many algorithms are proposed to tackle the problem of ED by learning-based approaches \cite{lin2020learning,jiang2020learning,wang2020event}, \ie,
\begin{equation}\label{eq:rednet}
    {\bf I}_{t} = \mbox{ED-Net} \left({\bf B},{\bf E}_T\right), \quad t\in T
\end{equation}
which are commonly trained over a synthetic dataset $\mathcal{D}_s \triangleq \{\hat{\bf B}_k,\hat{\bf E}_{T_k},\tilde{\mathcal{G}}_k\}_{k}$ where $\hat{\bf B}_k$ and $\hat{\bf E}_{T_k}$ are synthesized from the ground-truth sequence $\tilde{\mathcal{G}}_k$, \ie, $\hat{\bf E}_{T_k}=\mbox{ESIM}(\tilde{\mathcal{G}}_k)$ \cite{rebecq2018esim} and $\hat{\bf B}_k=\mbox{Avg}(\tilde{\mathcal{G}}_k)$.

\noindent{\bf Real-World Events}. Real-world events $\tilde{\bf E}_{T_k}$ are different from the simulated events $\hat{\bf E}_{T_k}$ in two aspects: (1) {\it event noises in spatial domain} are principally induced by physical intrinsic camera imperfections (\etc, variant event threshold), background activities, false negatives, \etc, \cite{gallego2020event}; (2) {\it event noises in temporal domain} are produced owing to the limited read-out bandwidth and bring timing variances \cite{inivaiton2020}, which commonly exist at a fast motion speed when the motion blur happens. Statistics of noises from both aspects are too complicated to be addressed in current event simulators \cite{rebecq2018esim}, leading to inconsistency between synthetic dataset $\mathcal{D}_s$ and its real-world counterparts. 

For spatial event noises, \cite{lin2020learning} has built a real event dataset $\mathcal{D}_e \triangleq \{\hat{\bf B}_k,\tilde{\bf E}_{T_k},\tilde{\mathcal{G}}_k\}_{k}$ composed of the real-world events $\tilde{\bf E}_{T_k}$, the real-world sharp clear ground-truth images $\tilde{\mathcal{G}}_k$ captured at an ultra slow speed to avoid motion blurs, and synthesized blurry images $\hat{\bf B}_k=\mbox{Avg}(\tilde{\mathcal{G}}_k)$. Theoretically, training on $\mathcal{D}_e$ is able to take into account the spatial noises, while the temporal noises cannot be considered since they always accompany with real-world motion blurs where no ground-truth images can be obtained. It raises the problem of learning on unlabeled real-world dataset $\mathcal{D}_r \triangleq \{\tilde{\bf B}_k,\tilde{\bf E}_{T_k}\}_{k}$, with real-world blurry images $\tilde{\bf B}_k$ and real-world events $\tilde{\bf E}_{T_k}$. 

\noindent{\bf Real-World Motions}. Motions and intensity textures are coupled together based on the photometric consistency and thus commonly exploited for the motion deblur \cite{jin2019learning,rengarajan2020photosequencing}. For simplicity, most existing motion deblur algorithms assume linear motion during the exposure time $T$ \cite{pan2020single,chen2018reblur2deblur}, where the motion flow stays constant. However, it is not always correct for dynamic scenes since real-world motions might be nonlinear especially at a fast motion speed. 


Thus the main obstacles of leveraging the power of events to learn motion deblurring in real world are in two folds. First, the network should be trained on real-world data with real events and real motion blurs. Second, the motion non-linearity should be considered to deal with motion blurs in real-world complex dynamic scenes.

\section{Method}

\begin{figure}[!t]
	\centering
	\includegraphics[width=\linewidth]{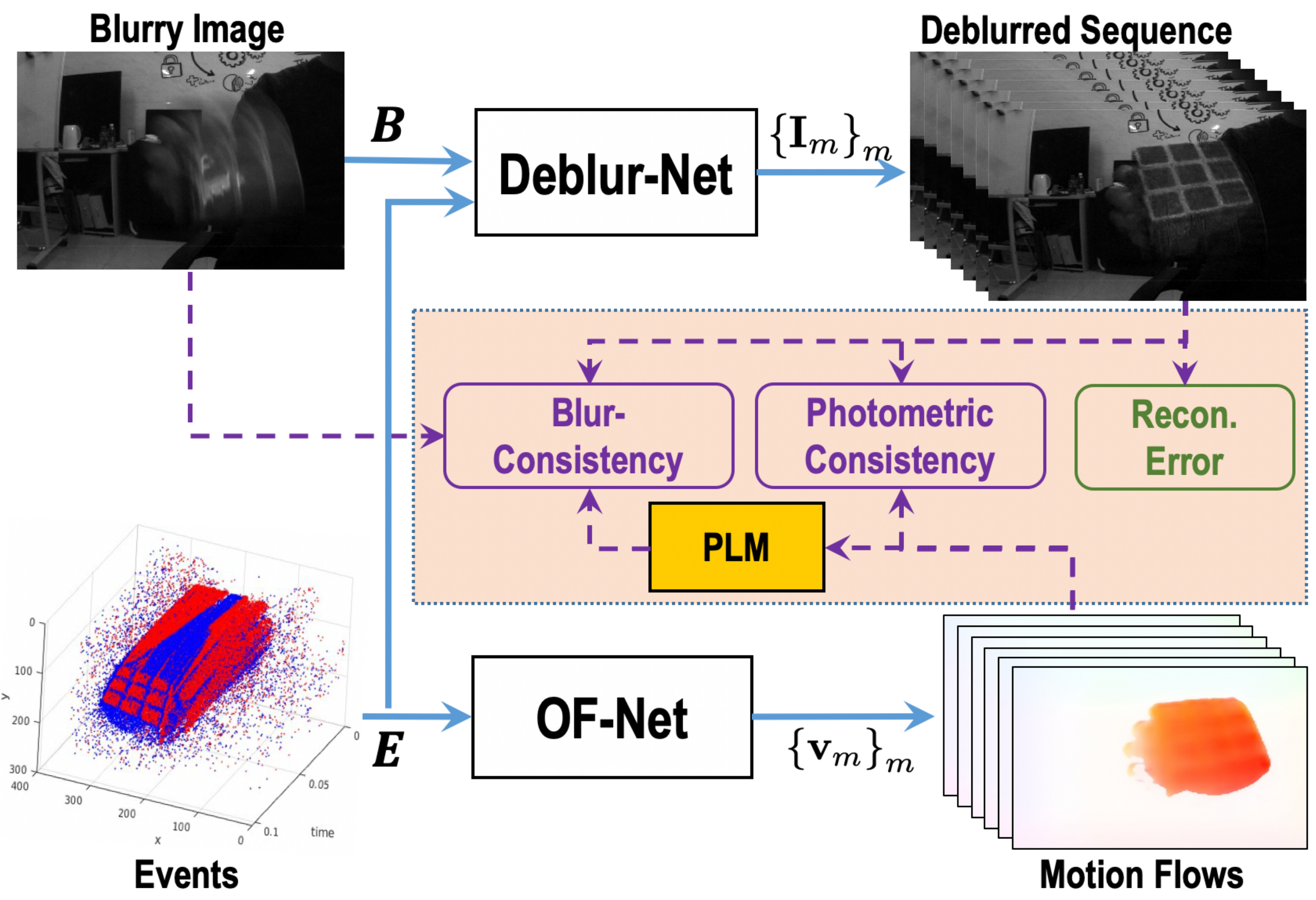} 
	\caption{Overview of the proposed learning framework for Real-world Event based motion Deblurring Network (RED-Net), where the blur-consistency and photometric-consistency provide self-supervised losses for real-world datasets and the reconstruction error provides the supervised loss for synthetic datasets.}
	\label{fig:Architecture}
\end{figure}

Existing event-based deblurring methods are usually developed within a supervised learning framework, of which performance is limited to the specific training data (simulated events or synthesized blurred image from the dataset captured with slow motion). Taking into account both deblurring and the physical image blur formation process, we construct a semi-supervised learning framework which can generalize well to real-world motion-blurred images caused by fast motion. 

Fig.~\ref{fig:Architecture} illustrates the overall architecture of our framework. The Deblur-Net takes a single motion-blurred image ${\mathbf B}$ and its event data ${\mathbf E}$ as input and outputs a sequence of sharp frames $\{\mathbf{I}_{m}\}_{m=0}^{M-1}$, where $M$ is the number of estimated sharp frames. The supervised loss that compares the estimated sharp frames with ground truth only works for labeled synthesized dataset. To guide the semi-supervised learning of Deblur-Net, our framework consists of an additional OF-Net for motion estimation, which takes event data ${\mathbf E}$ as input and outputs optical flow serving as motion information. On the strength of the physical blur formation process, a motion-blurred image can be re-rendered on the basis of estimated sharp images and motion information, which provides a \textit{blur-consistency} constraint by comparing with the original motion-blurred image. Besides, estimated sharp images are propagated by the estimated optical flow, which provides an additional constraint of \textit{photometric-consistency} by comparing the propagated sharp images with the directly recovered ones. 

Considering the motion non-linearity in real-world complex dynamic scenes, we take full advantage of high temporal resolution of event camera. Since event data is able to provide continuous motion information in a period, we do not need strong assumption on linear constant motion in a large time interval as other deblurring methods~\cite{gong2017motion, pan2020single, chen2018reblur2deblur, liu2020self}. Event data can be used in modeling highly nonlinear motion happened in the exposure time.

\subsection{Blur-consistency Constraint}
Providing the latent sharp image $\mathbf{L}_t(\mathbf{x})$ during the exposure time interval $T$, the blurring process can be physically formulated as the average of them in the duty cycle,
\begin{equation}
\label{eq:blur_model}
    \bar{\mathbf B}({\mathbf x}) =\frac{1}{|T|}\int_{t\in T} \mathbf{L}_t(\mathbf{x}) dt \approx \frac{1}{N}\sum_{n=0}^{N-1}\mathbf{L}_{n}(\mathbf{x}),
\end{equation}
where $\{\mathbf{L}_n\}_{n=0}^{N-1}$ is the discrete version of $\mathbf{L}_t$ and a large enough $N \gg M$ is required to achieve small discretization error. Providing recovered sequences $\{\mathbf{I}_{m}\}_{m=0}^{M-1}$ output from the Deblur-Net, one can resolve $\mathbf{L}_n$ by interpolating with $\{\mathbf{I}_{m}\}_{m=0}^{M-1}$ over the motion field ${\bf u}_n({\bf x})$ representing per-pixel flow from $\mathbf{L}_{0}$ to $\mathbf{L}_{n}$ \ie,
\begin{equation}
	\mathbf{L}_{n}(\mathbf{x})=\mathbf{L}_{0}(\mathbf{x}+\mathbf{u}_{n}(\mathbf{x})).
\end{equation}
For simplicity, a linear motion (LM) model is commonly assumed \cite{liu2020self,pan2020single,gong2017motion}, \ie,
\begin{equation} \label{eq:LM}
    \mathbf{u}_{n}(\mathbf{x}) = n \mathbf{v}(\mathbf{x}),
\end{equation}
with $\mathbf{v}$ denoting the per-pixel unit motion flow which is assumed to be constant during the exposure time interval $T$.

To consider the motion non-linearity for real-world scenario, we propose a piece-wise linear motion (PLM) model by making full use of rich motion information encoded in events. We divide event stream within the exposure time $T$ into $M-1$ equal time intervals and the LM model is assumed within each interval. Finally, we can get the following PLM model, for $n\in [mK,(m+1)K)$,
\begin{equation}\label{eq:plm}
    \mathbf{u}_{n}(\mathbf{x}) =  (n-mK)\mathbf{v}_m(\mathbf{x}) + \mathbf{u}_{mK}(\mathbf{x}), 
\end{equation}
with $\mathbf{v}_m$ the per-pixel unit motion flow that associates with the $m$-th time interval, $m=0,1,...,M-1$ and $K\triangleq N/M$ ($N$ can be chosen as integer times of $M$). And moreover,  $\mathbf{v}_m$ can be predicted from the OF-Net using events in the $m$-th time interval and resolve a nonlinear motion field $\mathbf{u}_{n}(\mathbf{x})$.

Then providing the sequence of sharp frames $\{\mathbf{I}_{m}\}_{m=0}^{M-1}$ output from the Deblur-Net, the $n$-th latent image $\mathbf{L}_n$ with $n\in [mK,(m+1)K)$ can be resolved by warping the $m$-th frame $\mathbf{I}_{m}$ using the motion field defined in \eqref{eq:plm}, 
\begin{equation} 
\begin{split}
    \mathbf{L}_n(\mathbf{x}) &= \mbox{Warp}\left(\mathbf{I}_{m}(\mathbf{x}),\mathbf{u}_{n}(\mathbf{x})\right) \\ & \triangleq  \mathbf{I}_m(\mathbf{x} + (n-mK)\mathbf{v}_m(\mathbf{x})).
\end{split}
\end{equation}

Hence, we can re-render a corresponding reblurred image based on \eqref{eq:blur_model} using the generated latent sharp images $\{\mathbf{L}_n\}_{n=0}^{N-1}$. A \textit{blur-consistency loss} is constructed between the reblurred image $\bar{\mathbf{B}}$ and the original blurry input ${\mathbf{B}}$:
\begin{equation}\label{eq:loss_blur}
	\mathcal{L}_{blur}=\|\bar{\mathbf{B}}-\mathbf{B}\|_{1}.
\end{equation}
By imposing this loss in the whole framework, we build a bridge between the motion-blurred image and its corresponding sharp clear images, which provides a self-supervision to the network. For real-world motion blurs with motion non-linearities, PLM is important to ensure the model accuracy and thus provide faithful supervision even with noise disturbances.

\subsection{Photometric-consistency Constraint}
In addition to the \textit{blur-consistency} constraint on the sequence level, further \textit{photometric-consistency} constraint is imposed, which utilizes the inter-frame connection between the temporal consistency of estimated sharp images and motion information encodes in events. We warp $\mathbf{I}_{m+1}$ to $\mathbf{I}_{m}$ with the predicted optical flow $\mathbf{v}_{m}$ via the OF-Net by feeding events in the $m$-th time interval and obtain the warped image:
\begin{equation}
	\bar{\mathbf{I}}_{m}(\mathbf{x})=\mathbf{I}_{m+1}(\mathbf{x}+K\mathbf{v}_{m}(\mathbf{x})).
\end{equation}
Then \textit{photometric-consistency} loss is constructed by comparing the propagated sharp
images with the recovered ones directly predicted by Deblur-Net:
\begin{equation}\label{eq:loss_photo}
	\mathcal{L}_{photo}=\frac{1}{M-1}\sum_{m=0}^{M-2}\|\bar{\mathbf{I}}_{m}-\mathbf{I}_{m}\|_{1}.
\end{equation}

\subsection{Optimization}
 We propose to train the Deblur-Net and OF-Net over a partially labeled dataset, composed of the synthetic dataset $\mathcal{D}_s$ with ground-truth sharp clear images and the real-world dataset $\mathcal{D}_r$ without ground truth. Apparently, the blur-consistency and the photometric-consistency are applicable for both $\mathcal{D}_s$ and $\mathcal{D}_r$, and provide self-supervised losses respectively defined as $\mathcal{L}_{blur}^s,\mathcal{L}_{photo}^s$ and $\mathcal{L}_{blur}^r,\mathcal{L}_{photo}^r$ according to \eqref{eq:loss_blur} and \eqref{eq:loss_photo}. Besides, for the synthetic dataset $\mathcal{D}_s$ with a sequence of ground-truth sharp clear images $\tilde{\mathcal{G}}$, we use it to supervise the network via a reconstruction error loss,
 \begin{equation}\label{eq:loss_error}
	\mathcal{L}_{error}=\frac{1}{M}\sum_{m=0}^{M-1}||\mathbf{I}_{m}-\mathbf{G}_m||_{1},
\end{equation}
 where $\mathbf{G}_{m} \in \tilde{\mathcal{G}}$.
 Thus the overall function is as follows:
\begin{equation}
	\mathcal{L}=\mathcal{L}_{error}^{s}+\alpha\mathcal{L}_{blur}^{s}+\beta\mathcal{L}_{photo}^{s}+\gamma\mathcal{L}_{blur}^{r} + \delta\mathcal{L}_{photo}^{r},
\end{equation}
with $\alpha$, $\beta$, $\gamma$ and $\delta$ denoting the balancing parameters. 

\section{Experiments}
\subsection{Experimental Settings}
\noindent{\bf Dataset.} The proposed Real-world Event-based motion Deblurring network (\myname-Net) is trained in a self-supervised manner, where one synthetic dataset ({\it GoPro}) is provided for training with ground-truth and two datasets respectively captured by a DAVIS240C camera ({\it HQF} provided in \cite{stoffregen2020reducing} containing real events but blur-free intensity frames) and a DAVIS346 camera ({\it RBE} built in this paper for the real-world scenario with real events and real motion blurs).

\textit{GoPro}: Based on the GoPro dataset \cite{nah2019ntire}, we build a synthetic dataset as \cite{wang2020event} composed of simulated events and synthetic blurry images as well as sharp clear ground-truth images. We first increase the frame rate by interpolating 7 images between consecutive frames~\cite{niklaus2017video} and then generate both events and blurry images based on the interpolated high frame-rate sequences. ESIM~\cite{rebecq2018esim} is exploited to simulate events with consideration of per-pixel threshold variation. And blurry images are simply obtained by averaging over 49 consecutive images. 

\textit{HQF}: We construct a similar dataset as Blur-DVS \cite{lin2020learning} based on HQF dataset~\cite{stoffregen2020reducing}, which contains real-world events and sharp clear ground-truth frames captured simultaneously from a DAVIS240C that are well-exposed and minimally motion-blurred. Providing the ground-truth frames, motion blurs can be synthesized using the same manner as the GoPro dataset. Finally, the HQF dataset contains real-world events and synthesized blurry images as well as the ground-truth frames. The ground-truth frames are provided only for quantitative evaluation.
In training stage of our proposed RED-Net over the HQF dataset, only real events and synthesized blurry images are used.

\textit{RBE}: A Real-world Blurry images and Events (RBE) dataset is built with a DAVIS346 camera and only contains real-world events and real-world motion blurs, which can be collected in a facilitated manner. Thus the RBE dataset can provide a large number of paired real-world data, which can be fed into the RED-Net and improve the adaptivity to the real-world scenario.

\noindent{\bf Implementation Details}
For the Deblur-Net and OF-Net modules, we take advantage of existing neural network architectures which have performed well in the past for the respective learning tasks. Inspired by that a sharp image can be mapped from a blurry image using a residual term encoded with events~\cite{pan2019bringing}, we select a residual network to predict a sequence of sharp frames from a single motion-blurred image and its event data. In particular, we adopt the residual network from Jin \textit{et al.}~\cite{jin2019learning}. For the OF-Net, we adopt the EV-FlowNet architecture from Zhu \textit{et al.}~\cite{zhu2018ev}, which is widely used to predict optical flow from events.

Our network is implemented using Pytorch on a single NVIDIA Geforce RTX 3090 GPU. During training, we randomly crop the samples into $128\times 128$ patches. Adam optimizer is used and the maximum epoch of training iterations is set to 30. The learning rate starts at $10^{-4}$, then decays by $25\%$ every five epochs from the $15$-th epoch. The weighting factors $\alpha, \beta, \gamma$ and $\delta$ are all set to $1$, the number of deblurred images $\{\mathbf{I}_{m}\}_{m=0}^{M-1}$ and latent sharp images $\mathbf{L}_{n}$ warped from $\mathbf{I}_{m}$ are respectively $M=7$ and $K=11$. The optical flow module is pretrained on the Multi-Vehicle Stereo Event Camera dataset (MVSEC)~\cite{zhu2018ev}, where the reference ground-truth optical flow is provided.

\noindent{\bf \myname-Nets.} To validate the effectiveness of exploiting real-world data, \myname-Net is trained respectively over three different datasets, \ie, GoPro, GoPro+HQF and GoPro+RBE, and final networks are respectively denoted as {\it \myname-GoPro, \myname-HQF and \myname-RBE}. Specifically, \myname-GoPro is trained only over synthesized GoPro dataset with the supervision of ground-truth frames. Both \myname-HQF and \myname-RBE are trained in a self-supervised manner, where \myname-HQF uses real-world events but synthetic motion blurs, while \myname-RBE uses real-world events and motion blurs. 

To validate the superiority of piece-wise linear motion model (PLM), we replace it with the linear motion model (LM) as \eqref{eq:LM}, and train \myname-Net respectively on GoPro and GoPro+RBE datasets. Specifically, the resulted network is denoted as { \myname-(GoPro/RBE)-LM} and its counterpart with PLM is denoted as { \myname-(GoPro/RBE)-PLM}. By default, \myname-(GoPro/RBE) means \myname-(GoPro/RBE)-PLM in the following sections.

\subsection{Results of Optical Flow}\label{sec:of}
\begin{figure}[!htb]
	\centering
	\subfigure[]{
		\begin{minipage}[t]{0.24\linewidth}
			\centering\label{fig:flow_reslut1}
			\includegraphics[width=0.98\linewidth]{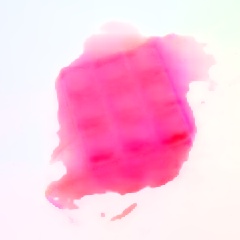}
		\end{minipage}%
	}\hspace{-1.5mm}
	\subfigure[]{
		\begin{minipage}[t]{0.24\linewidth}
			\centering\label{fig:flow_reslut4_linear}
			\includegraphics[width=0.98\linewidth]{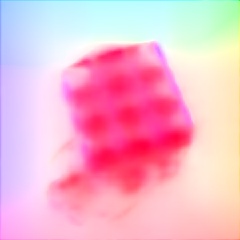}
		\end{minipage}%
	}\hspace{-1.5mm}
	\subfigure[]{
		\begin{minipage}[t]{0.24\linewidth}
			\centering\label{fig:flow_reslut2}
			\includegraphics[width=0.98\linewidth]{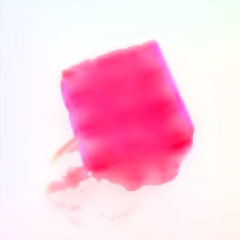}
		\end{minipage}%
	}\hspace{-1.5mm}
	\subfigure[]{
		\begin{minipage}[t]{0.24\linewidth}
			\centering\label{fig:flow_reslut3}
			\includegraphics[width=0.98\linewidth]{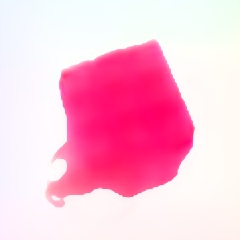}
		\end{minipage}%
	}\vspace{-1.5mm}\\
	\subfigure[]{
		\begin{minipage}[t]{0.24\linewidth}
			\centering\label{fig:flow_reslut4}
			\includegraphics[width=0.98\linewidth]{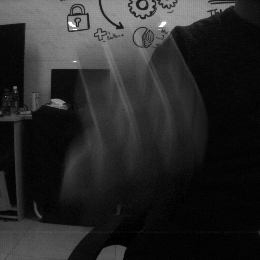}
		\end{minipage}%
	}\hspace{-1.5mm}
	\subfigure[]{
		\begin{minipage}[t]{0.24\linewidth}
			\centering\label{fig:flow_reslut4_linear_recon}
			\includegraphics[width=0.98\linewidth]{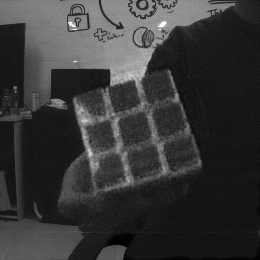}
		\end{minipage}%
	}\hspace{-1.5mm}
	\subfigure[]{
		\begin{minipage}[t]{0.24\linewidth}
			\centering
			\label{fig:flow_reslut5}
			\includegraphics[width=0.98\linewidth]{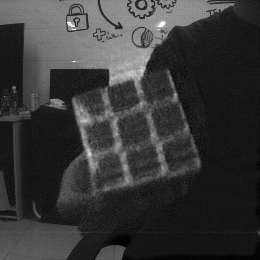}
		\end{minipage}%
	}\hspace{-1.5mm}
	\subfigure[]{
		\begin{minipage}[t]{0.24\linewidth}
			\centering\label{fig:flow_reslut6}
			\includegraphics[width=0.98\linewidth]{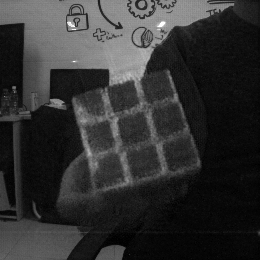}
		\end{minipage}%
	}
	\caption{Optical flow output of OF-Net: (a) pre-trained on MVSEC (equivalent to EV-FlowNet), (b) \myname-GoPro-LM, (c) \myname-GoPro-PLM and (d) \myname-RBE. The deblurred results (f), (g) and (h) of the motion blurred image (e) are respectively corresponding to (b), (c) and (d).}	
	\label{fig:flow_reslut}
\end{figure}

Fig.~\ref{fig:flow_reslut} presents the optical flow (OF) outputs of OF-Net from different \myname-Nets and their corresponding deblur results of a real-world motion blurred image containing a fast moving {\it magic cube}. For comparison, the output of OF-Net pretrained on MVSEC dataset (\ie EV-FlowNet) is also given. Apparently, the output of OF-Net from \myname-RBE trained on real data achieves the best performance with consistent motion and sharp clear edge. 
Thus the deblur image of \myname-RBE has the best qualitative appearance as well.

To reveal the advantage of PLM, we compare the outputs of OF-Net from \myname-GoPro-LM as shown in Fig.~\ref{fig:flow_reslut4_linear} and \myname-GoPro-PLM as shown in Fig.~\ref{fig:flow_reslut2}. It is obvious that \myname-GoPro-PLM gives better OF result with clear edges and stable background while \myname-GoPro-LM suffers from blurry effects and background movement. Furthermore, when violating the linear motion assumption, the result of OF-Net from \myname-GoPro-LM is even worse than its origin, i.e. EvFlowNet as shown in Fig.~\ref{fig:flow_reslut1} and \ref{fig:flow_reslut4_linear}.

Besides, we compare the OF results of EvFlowNet and two OF-Nets respectively from \myname-GoPro-PLM and \myname-RBE as shown in Fig.~\ref{fig:flow_reslut1}, \ref{fig:flow_reslut2} and \ref{fig:flow_reslut3}. With the supervision of blurry image and the intensity sequence according to the blur-consistency loss \eqref{eq:loss_blur}, OF-Nets from both \myname-GoPro-PLM and \myname-RBE can give better results than EvFlowNet. It reveals that the motion information embedded in the blurry images helps the prediction of OF within our framework.  Moreover, OF-Net from \myname-RBE trained over the real dataset outperforms that of \myname-GoPro-PLM trained over the synthetic dataset and gives more consistent motions of the cube surface as shown in Fig.~\ref{fig:flow_reslut2} and \ref{fig:flow_reslut3}. And it validates the advantage and necessity of training on real-world dataset.

\begin{table*}[]
\small
\centering
\caption{Quantitative comparisons of proposed \myname-Nets trained over different datasets to the state-of-the-arts. \myname-GoPro, \myname-HQF, and \myname-RBE are respectively trained on GoPro, GoPro+HQF and GoPro+RBE. Note that LEDVDI only outputs 6 frames for sequence prediction, while the others output 7 frames. }
\begin{tabular}{llllllllllll}
\hline
\multirow{3}{*}{Method}                                    & \multicolumn{5}{c}{Single frame prediction comparison}                                                                                                   &  & \multicolumn{5}{c}{Sequence prediction comparison}                                                                                                       \\ \cline{2-6} \cline{8-12} 
                                                           & \multicolumn{2}{c}{GoPro}                                                 &  & \multicolumn{2}{c}{HQF}                                                   &  & \multicolumn{2}{c}{GoPro}                                                 &  & \multicolumn{2}{c}{HQF}                                                   \\ \cline{2-3} \cline{5-6} \cline{8-9} \cline{11-12} 
                                                           & PSNR                            & SSIM                            &  & PSNR                            & SSIM                            &  & PSNR                            & SSIM                            &  & PSNR                            & SSIM                            \\ \hline
blur2mflow~\cite{gong2017motion} & 21.818                          &  0.6454                         &  & 21.539                          &  0.6298                         &  & /                               & /                               &  & /                               & /                               \\
LEVS~\cite{jin2018learning}      & 21.950                          & 0.6406                          &  & 21.900                          & 0.6367                          &  & 19.893                          & 0.5546                          &  & 19.068                          & 0.5403                          \\
EDI~\cite{pan2019bringing}       & 21.497                          & 0.6510                          &  & 20.321                          & 0.6212                          &  & 20.945                          & 0.6326                          &  & 19.081                          & 0.5873                          \\
eSL-Net~\cite{wang2020event}     & 24.791                          & 0.8009                          &  &  20.438                         & 0.6017                          &  & 23.955                          & 0.7578                          &  &  19.866                         & 0.5851                          \\
LEDVDI~\cite{lin2020learning}    & 22.856                          &  0.7334                         &  & 22.221                          & 0.7567                          &  & 22.673                              & 0.7329                              &  & 21.558                              & {\bf 0.7355}                              \\ \hline
 \myname-GoPro                          & {\bf 28.984}       & {\bf 0.8499}       &  & 24.149                          & 0.7332                          &  & {\bf 28.343}       & {\bf 0.8359}       &  & 23.118                          & 0.7116                          \\
 \myname-HQF                            & \underline{27.137} & {0.8361} &  & \underline{25.650}       & \bf{0.7661} &  & \underline{26.640} &  0.8204                         &  & \underline{23.872}       & {0.7292}       \\
 \myname-RBE                            & 26.672                          & \underline{0.8372}                          &  & \bf{25.717} & \underline{0.7629}                         &  &  26.302                         & \underline{0.8247} &  & \bf{24.076} & \underline{0.7340} \\ \hline
\end{tabular}
\label{table1}
\end{table*}

\subsection{Comparisons with State-of-the-art Methods}
Since our method is able to remove motion blur and reconstruct a sequence of latent sharp clear images, the deblur performance is evaluated on both single frame and sequence reconstructions.  We compare the proposed \myname-Nets to state-of-the-art conventional deblurring methods including blur2mflow~\cite{gong2017motion} and LEVS~\cite{jin2018learning},  and event-based motion deblurring methods including EDI~\cite{pan2019bringing}, eSL-Net~\cite{wang2020event} and LEDVDI~\cite{lin2020learning}. Due to the lack of ground-truth for real-world motion blurs, we exploit the HQF dataset constructed from blur-free frames for quantitative evaluations, where the effectiveness of training with real-world events are validated.
The quantitative results are presented in Tab.\ref{table1} and show that 
the proposed \myname-Nets bring remarkable improvements comparing to the state-of-the-arts, on average {3.5 dB} gain on single frame prediction and {2.5 dB} gain on sequence prediction over the HQF dataset with real-world events. Detailed analyses are presented in the following.

\noindent{\bf Inconsistency of Simulated and Real-world Events.} We first compare the deblurring performance of conventional methods, \ie, blur2mflow and LEVS to  event-based methods, \ie, EDI and eSL-Net, over the synthetic GoPro dataset and the HQF dataset with real-world events. It is shown that EDI provides comparable performance to learning based conventional methods with the help of events. For the learning based approaches, eSL-Net enhanced with simulated events outperforms blur2mflow and LEVS by a large margin on the synthetic GoPro dataset. However, the overwhelming performance of eSL-Net is not retained on the HQF dataset with real-world events, which reflects the inconsistency of simulated and real-world events.

\noindent{\bf Effectiveness of Training with Real-world Events.} Comparing to RED-GoPro trained only on synthetic dataset, the other two RED-Nets, \ie, RED-HQF and RED-RBE, trained respectively over the HQF and RBE dataset with real events, bring a performance drop on the synthetic GoPro dataset but achieve the PSNR gain on the HQF dataset with real-world events. Specifically, RED-HQF obtains on average 1.1 dB gain on PSNR, which shows the effectiveness of training with real-world events and bridges the sim-to-real gap. Furthermore, we validate this point by training RED-RBE with the real-world data in the RBE dataset which can be constructed in a facilitated manner. 

\noindent{\bf Event-based Motion Deblur with Real-World Events.} As shown in Tab.~\ref{table1}, \myname-HQF/RBE outperforms EDI and eSL-Net over the HQF dataset with real-world events by a large margin. Furthermore, we exploit the recently proposed network, \ie, LEDVDI trained with real-world events for fair comparisons. Different from our proposed method, LEDVDI is trained with real-world events in a supervised manner. Our approach requires neither blurry synthesis nor strict slow moving speed to collect blur-free ground-truth images, and thus is much easier to be adopted in real-world scenario. From the quantitative results in Tab.~\ref{table1}, it is shown that both \myname-HQF and \myname-RBF achieve better performance than LEDVDI for single frame prediction and sequence prediction, which validates the superiority of our proposed framework. Note that LEDVDI deals with single frame prediction and sequence prediction separately with different networks, while our proposed \myname-HQF/RBE tackle both problems in one network.

\noindent{\bf Real-World Motion Deblur.} The above quantitative comparisons are all based on synthesized motion blurs. Due to the lack of ground-truth frames, the performance with real-world motion blurs are qualitatively investigated, as shown in Fig.~\ref{fig:blur_result}. We choose two scenes captured manually with a DAVIS346 camera, \ie, {\it labfloor} and {\it window}. Without the aid of events, blur2mflow and LEVS fail to recover sharp clear images. For event-based methods, EDI fails to cope with severe motion blurs as illustrated in {\it window}. Among the learning based approaches, RED-RBE achieves the best visualization performance, while the others are likely to give over smoothed effects or generate {\it halo artifacts} around black edges, for instance the bottom of garbage can in {\it labfloor} and the top-left corner of TV in {\it window}. 
Compared with RED-HQF and LEDVDI trained with synthesized motion blurs, RED-RBE is trained with real-world motion blurs which can alleviate the problem of event noise in temporal domain, \eg, {\it timing chaos} caused by limited read-out bandwidth and thus give less halo artifacts.

\def\ssxxs{(0.5,-1.1)} 
\def\ssyys{(-0.45,0.7)}

\def\ssxx{(-0.7,-0.5)} 
\def\ssyy{(-0.45,0.7)}
\def\ssizz{1.5cm}
\def\sswidth{0.245\textwidth}
\def\ssmag{3}
\def\scc{(2.12,1.4)}
\begin{figure*}
	\centering
	        \begin{tikzpicture}[spy using outlines={green,magnification=\ssmag,size=\ssizz},inner sep=0]
			\node {\includegraphics[width=\sswidth]{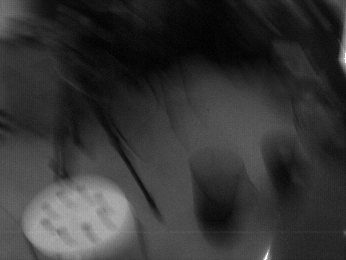}};
			\spy on \ssxxs in node [left] at \ssyys;
			\node [anchor=east] at \scc {\textcolor{white}{\bf Blurry Image}};
			\end{tikzpicture}
			\begin{tikzpicture}[spy using outlines={green,magnification=\ssmag,size=\ssizz},inner sep=0]
			\node {\includegraphics[width=\sswidth]{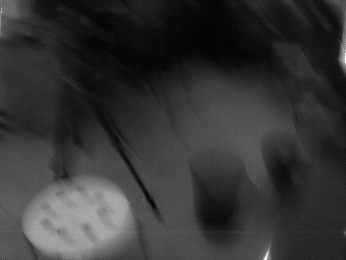}};
			\spy on \ssxxs in node [left] at \ssyys;
			\node [anchor=east] at \scc {\textcolor{white}{\bf blur2mflow}};
			\end{tikzpicture}
			\begin{tikzpicture}[spy using outlines={green,magnification=\ssmag,size=\ssizz},inner sep=0]
			\node {\includegraphics[width=\sswidth,trim={0 0 0 5px},clip=true]{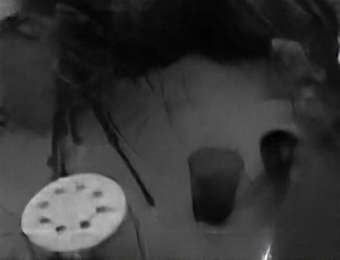}};
			\spy on \ssxxs in node [left] at \ssyys;
			\node [anchor=east] at \scc {\textcolor{white}{\bf LEVS}};
			\end{tikzpicture}
			\begin{tikzpicture}[spy using outlines={green,magnification=\ssmag,size=\ssizz},inner sep=0]
			\node {\includegraphics[width=\sswidth]{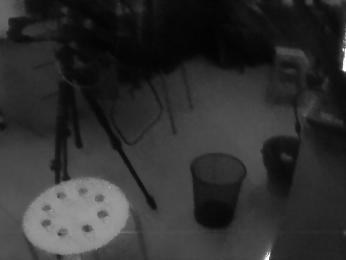}};
			\spy on \ssxxs in node [left] at \ssyys;
			\node [anchor=east] at \scc {\textcolor{white}{\bf EDI}};
			\end{tikzpicture}
			\\
			\begin{tikzpicture}[spy using outlines={green,magnification=\ssmag,size=\ssizz},inner sep=0]
			\node {\includegraphics[width=\sswidth]{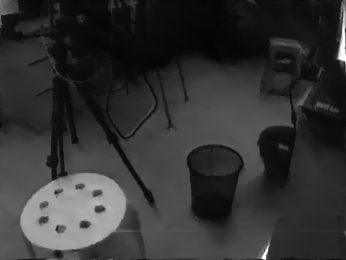}};
			\spy on \ssxxs in node [left] at \ssyys;
			\node [anchor=east] at \scc {\textcolor{white}{\bf eSL-Net}};
			\end{tikzpicture}
			\begin{tikzpicture}[spy using outlines={green,magnification=\ssmag,size=\ssizz},inner sep=0]
			\node {\includegraphics[width=\sswidth]{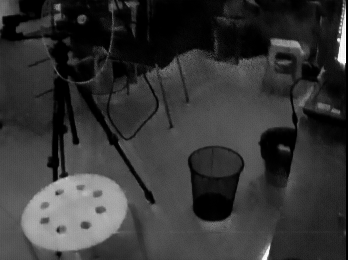}};
			\spy on \ssxxs in node [left] at \ssyys;
			\node [anchor=east] at \scc {\textcolor{white}{\bf LEDVDI}};
			\end{tikzpicture}
			\begin{tikzpicture}[spy using outlines={green,magnification=\ssmag,size=\ssizz},inner sep=0]
			\node {\includegraphics[width=\sswidth]{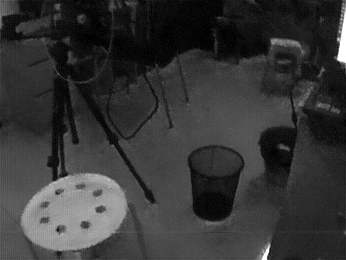}};
			\spy on \ssxxs in node [left] at \ssyys;
			\node [anchor=east] at \scc {\textcolor{white}{\bf RED-HQF}};
			\end{tikzpicture}
			\begin{tikzpicture}[spy using outlines={green,magnification=\ssmag,size=\ssizz},inner sep=0]
			\node {\includegraphics[width=\sswidth]{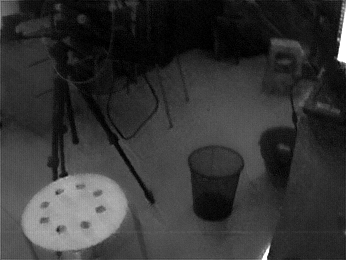}};
			\spy on \ssxxs in node [left] at \ssyys;
			\node [anchor=east] at \scc {\textcolor{white}{\bf RED-RBF}};
			\end{tikzpicture}
			\begin{tikzpicture}[spy using outlines={green,magnification=\ssmag,size=\ssizz},inner sep=0]
			\node {\includegraphics[width=\sswidth]{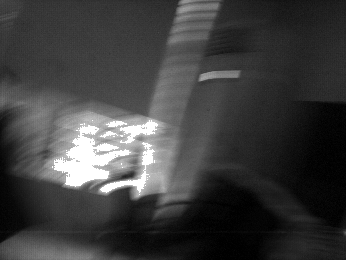}};
			\spy on \ssxx in node [left] at \ssyy;
			\node [anchor=east] at \scc {\textcolor{white}{\bf Blurry Image}};
			\end{tikzpicture}
			\begin{tikzpicture}[spy using outlines={green,magnification=\ssmag,size=\ssizz},inner sep=0]
			\node {\includegraphics[width=\sswidth]{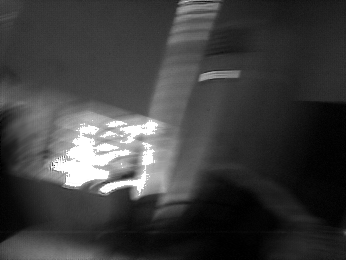}};
			\spy on \ssxx in node [left] at \ssyy;
			\node [anchor=east] at \scc {\textcolor{white}{\bf blur2mflow}};
			\end{tikzpicture}
			\begin{tikzpicture}[spy using outlines={green,magnification=\ssmag,size=\ssizz},inner sep=0]
			\node {\includegraphics[width=\sswidth,trim={0 0 0 5px},clip=true]{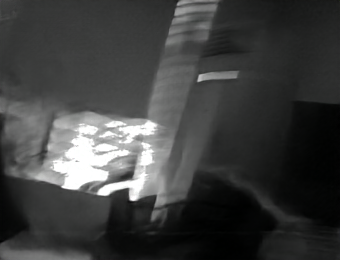}};
			\spy on \ssxx in node [left] at \ssyy;
			\node [anchor=east] at \scc {\textcolor{white}{\bf LEVS}};
			\end{tikzpicture}
			\begin{tikzpicture}[spy using outlines={green,magnification=\ssmag,size=\ssizz},inner sep=0]
			\node {\includegraphics[width=\sswidth]{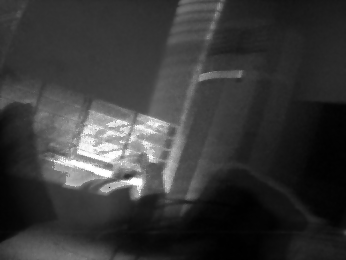}};
			\spy on \ssxx in node [left] at \ssyy;
			\node [anchor=east] at \scc {\textcolor{white}{\bf EDI}};
			\end{tikzpicture}
			\\
			\begin{tikzpicture}[spy using outlines={green,magnification=\ssmag,size=\ssizz},inner sep=0]
			\node {\includegraphics[width=\sswidth]{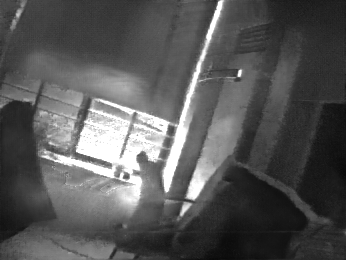}};
			\spy on \ssxx in node [left] at \ssyy;
			\node [anchor=east] at \scc {\textcolor{white}{\bf eSL-Net}};
			\end{tikzpicture}
			\begin{tikzpicture}[spy using outlines={green,magnification=\ssmag,size=\ssizz},inner sep=0]
			\node {\includegraphics[width=\sswidth]{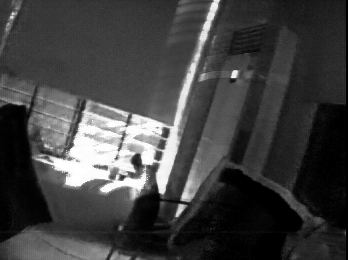}};
			\spy on \ssxx in node [left] at \ssyy;
			\node [anchor=east] at \scc {\textcolor{white}{\bf LEDVDI}};
			\end{tikzpicture}
			\begin{tikzpicture}[spy using outlines={green,magnification=\ssmag,size=\ssizz},inner sep=0]
			\node {\includegraphics[width=\sswidth]{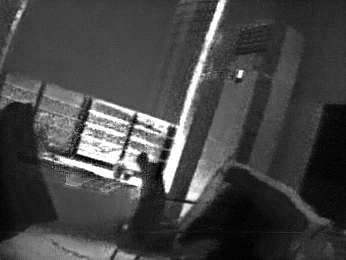}};
			\spy on \ssxx in node [left] at \ssyy;
			\node [anchor=east] at \scc {\textcolor{white}{\bf RED-HQF}};
			\end{tikzpicture}
			\begin{tikzpicture}[spy using outlines={green,magnification=\ssmag,size=\ssizz},inner sep=0]
			\node {\includegraphics[width=\sswidth]{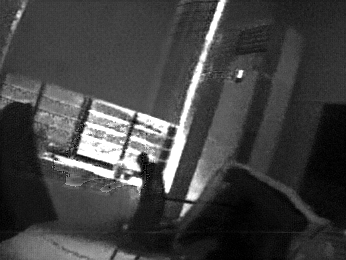}};
			\spy on \ssxx in node [left] at \ssyy;
			\node [anchor=east] at \scc {\textcolor{white}{\bf RED-RBF}};
			\end{tikzpicture}

			\caption{Qualitative results of motion deblur for 2 blurry scenes by 7 different methods where the top-two rows correspond to {\it labfloor} and  the bottom-two correspond to {\it window}. For each scene, from top-left to bottom-right are respectively the blurry image and its deblurred result by blur2mflow \cite{gong2017motion}, LEVS \cite{jin2018learning}, EDI \cite{pan2019bringing}, eSL-Net \cite{wang2020event}, LEDVDI \cite{lin2020learning} and our proposed RED-HQF and RED-RBE.}	
\label{fig:blur_result}
\end{figure*}
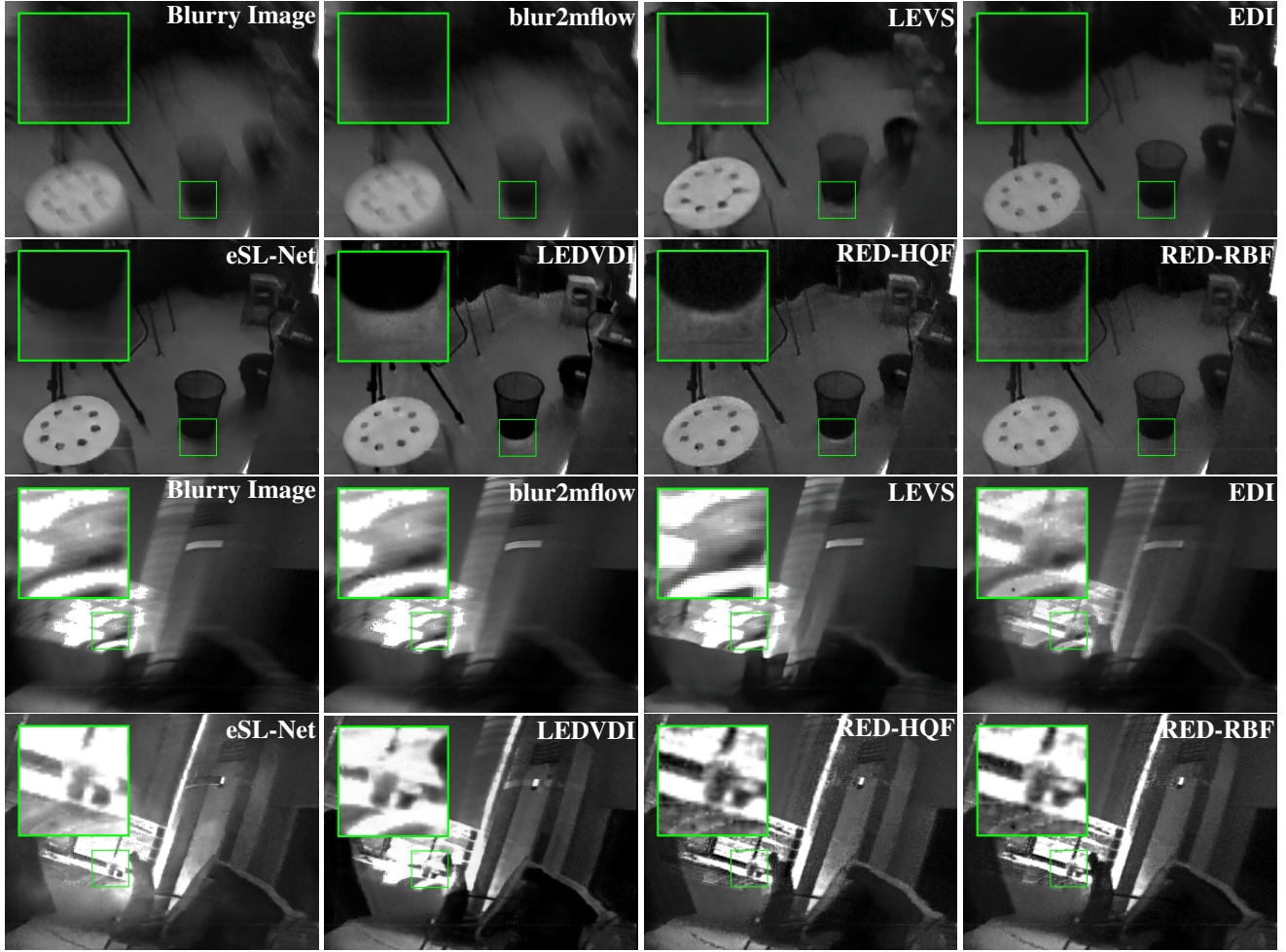

\begin{table}[!htb]
\small
\centering
\caption{Ablation study of proposed framework w/o {\it synthesized events} (SynEv), {\it blurring with LM} (LM) , {\it blurring with PLM} (PLM) and {\it real-world events} (ReEv). }
\label{tab:table2}
\resizebox{.47\textwidth}{!}{%
\begin{tabular}{l|llll|ll}
\hline
Methods&SynEv & LM & PLM & ReEv & {GoPro} & {HQF} \\ \hline
Deblur-Net \cite{jin2019learning} &            &              &              &              & 23.563 & 21.373       \\
Deblur-Net-GoPro&$\checkmark$ &              &              &              & 28.451 & 23.553     \\
RED-GoPro-LM&$\checkmark$ & $\checkmark$ &              &              & 28.752 & 23.988    \\
RED-GoPro&$\checkmark$ &  & $\checkmark$             &              & \bf 28.984 & 24.149      \\
RED-RBE-LM&$\checkmark$ & $\checkmark$ &  &$\checkmark$              & 26.451 & 25.266      \\
RED-RBE&$\checkmark$ &  & $\checkmark$ &$\checkmark$              &  26.672 & \bf 25.717   \\ \hline
\end{tabular}%
}
\vspace{-1em}
\end{table}

\subsection{Ablation Study}
\label{sec:Ablation}
The proposed \myname-Net improves the performance of motion deblurring by self-supervising with real-word events. To achieve this end, optical flows are predicted from events with which the piece-wise linear motion model (PLM) is proposed to accurately model the blurry procedure even with motion non-linearities. To find out what contributes to the superior performance of the approach, we compare a few variants with and without use of synthesized events (SynEv), blurring with LM (LM), blurring with PLM (PLM) and real-world events (ReEv), as shown in Tab.~\ref{tab:table2}. From the table, we can draw the following conclusions:

\noindent{\bf Importance of Events.} We validate the importance of events by training the Deblur-Net without events over the synthetic GoPro dataset, and the quantitative deblur results over 2 dataset are shown in the first row of Tab.~\ref{tab:table2}. Comparing to \myname-Nets with events, there is a large performance gap. It shows that the rich motion information encoded in events can effectively improve the deblurring performance.

\noindent{\bf Linear \vs piece-wise Linear Motion Model.} 
With the optical flow output of the OF-Net, motions provide the inter-frame relationship inside the reconstructed sequence as well as the blurry consistency which relies on the physical blurry procedure. Inaccurate motion model will violate the blurry consistency and thus degrades the deblurring performance. We validate this point by respectively apply LM and PLM in \eqref{eq:blur_model} and their performance can be referred from Tab.~\ref{tab:table2} in the 3rd, 4th, 5th and 6th rows. Apparently, \myname-Nets with PLM outperform that with LM and achieve PSNR gains up to 0.5 dB on the real-world dataset after training with real-world events, while it only gains 0.16 dB without real-world events.
It demonstrates that PLM plays an important role especially for real-world events.

\noindent{\bf Synthetic \vs Real-world Events.} To validate this point, we train \myname-Net respectively on the synthetic GoPro dataset, \ie, \myname-GoPro and on the real-world RBE dataset, \ie, \myname-RBE. Their results are shown in the 4th and 6th rows of Tab.~\ref{tab:table2}. And we also trained \myname-GoPro and \myname-RBE with LM and present their results respectively in the 3rd and 5th rows of Tab.~\ref{tab:table2}. By comparisons, we can easily find that the self-supervision with real-world events can achieve PSNR gains 1.6 dB with PLM (1.3 dB with LM) on real-world dataset. It shows that the supervision of real-world events plays a major role in improving the deblurring performance on real-world dataset.
\vspace{-0.2cm}

\section{Conclusion}
{A self-supervised learning framework for event-based motion deblurring is proposed where real-world events and real-world motion blurs are exploited to alleviate the performance degradation caused by data inconsistency. Motion non-linearities are also considered through a PLM model to improve the accuracy of physical blur-consistency. Within the proposed learning framework, we evaluate RED-Net over different datasets and validate the effectiveness of the proposed PLM-based blur-consistency and photometric-consistency over the real-world dataset. With a blurry image and corresponding events, the proposed RED-Net can produce a sequence of sharp clear intensity images as well as motion flows between them. Extensive experiments demonstrate that the proposed method can achieve the state-of-the-art with real-world events.}

{\small
\bibliographystyle{ieee_fullname}
\bibliography{egbib}

\begin{thebibliography}{10}\itemsep=-1pt

\bibitem{baldwin2020event}
R Baldwin, Mohammed Almatrafi, Vijayan Asari, and Keigo Hirakawa.
\newblock Event probability mask (epm) and event denoising convolutional neural
  network (edncnn) for neuromorphic cameras.
\newblock In {\em CVPR}, pages 1701--1710, 2020.

\bibitem{benosman2013event}
Ryad Benosman, Charles Clercq, Xavier Lagorce, Sio-Hoi Ieng, and Chiara
  Bartolozzi.
\newblock Event-based visual flow.
\newblock {\em IEEE Transactions on Neural Networks and Learning Systems},
  25(2):407--417, 2013.

\bibitem{brandli2014b}
Christian Brandli, Raphael Berner, Minhao Yang, Shih-Chii Liu, and Tobi
  Delbruck.
\newblock A 240$\times$180 130 db 3 $\mu$s latency global shutter
  spatiotemporal vision sensor.
\newblock {\em IEEE Journal of Solid-State Circuits}, 49(10):2333--2341, 2014.

\bibitem{chen2018reblur2deblur}
Huaijin Chen, Jinwei Gu, Orazio Gallo, Ming-Yu Liu, Ashok Veeraraghavan, and
  Jan Kautz.
\newblock Reblur2deblur: Deblurring videos via self-supervised learning.
\newblock In {\em IEEE International Conference on Computational Photography
  (ICCP)}, pages 1--9, 2018.

\bibitem{fergus2006removing}
Rob Fergus, Barun Singh, Aaron Hertzmann, Sam~T Roweis, and William~T Freeman.
\newblock Removing camera shake from a single photograph.
\newblock In {\em ACM SIGGRAPH}, pages 787--794. 2006.

\bibitem{gallego2020event}
Guillermo Gallego, Tobi Delbruck, Garrick~Michael Orchard, Chiara Bartolozzi,
  Brian Taba, Andrea Censi, Stefan Leutenegger, Andrew Davison, Jorg Conradt,
  Kostas Daniilidis, and Davide Scaramuzza.
\newblock Event-based vision: A survey.
\newblock {\em IEEE Transactions on Pattern Analysis and Machine Intelligence},
  2020.

\bibitem{gong2017motion}
Dong Gong, Jie Yang, Lingqiao Liu, Yanning Zhang, Ian Reid, Chunhua Shen, Anton
  Van Den~Hengel, and Qinfeng Shi.
\newblock From motion blur to motion flow: A deep learning solution for
  removing heterogeneous motion blur.
\newblock In {\em CVPR}, pages 2319--2328, 2017.

\bibitem{hyun2015generalized}
Tae Hyun~Kim and Kyoung Mu~Lee.
\newblock Generalized video deblurring for dynamic scenes.
\newblock In {\em CVPR}, pages 5426--5434, 2015.

\bibitem{inivaiton2020}
IniVation.
\newblock {\em Understanding the Performance of Neuromorphic Event-Based Vision
  Sensors}.
\newblock IniVation, https://inivation.com/, 1 edition, 05 2020.

\bibitem{jiang2020learning}
Zhe Jiang, Yu Zhang, Dongqing Zou, Jimmy Ren, Jiancheng Lv, and Yebin Liu.
\newblock Learning event-based motion deblurring.
\newblock In {\em CVPR}, pages 3320--3329, 2020.

\bibitem{jin2019learning}
Meiguang Jin, Zhe Hu, and Paolo Favaro.
\newblock Learning to extract flawless slow motion from blurry videos.
\newblock In {\em CVPR}, pages 8112--8121, 2019.

\bibitem{jin2018learning}
Meiguang Jin, Givi Meishvili, and Paolo Favaro.
\newblock Learning to extract a video sequence from a single motion-blurred
  image.
\newblock In {\em CVPR}, pages 6334--6342, 2018.

\bibitem{krishnan2011blind}
Dilip Krishnan, Terence Tay, and Rob Fergus.
\newblock Blind deconvolution using a normalized sparsity measure.
\newblock In {\em CVPR}, pages 233--240, 2011.

\bibitem{lagorce2016hots}
Xavier Lagorce, Garrick Orchard, Francesco Galluppi, Bertram~E Shi, and Ryad~B
  Benosman.
\newblock Hots: a hierarchy of event-based time-surfaces for pattern
  recognition.
\newblock {\em IEEE Transactions on Pattern Analysis and Machine Intelligence},
  39(7):1346--1359, 2016.

\bibitem{lichtsteiner128Times1282008}
Patrick Lichtsteiner, Christoph Posch, and Tobi Delbruck.
\newblock A 128 $\times$ 128 120 {dB} 15 $\mu$s {Latency} {Asynchronous}
  {Temporal} {Contrast} {Vision} {Sensor}.
\newblock {\em IEEE Journal of Solid-State Circuits}, 43(2):566--576, 2008.

\bibitem{lin2020learning}
Songnan Lin, Jiawei Zhang, Jinshan Pan, Zhe Jiang, Dongqing Zou, Yongtian Wang,
  Jing Chen, and Jimmy Ren.
\newblock Learning event-driven video deblurring and interpolation.
\newblock In {\em ECCV}, 2020.

\bibitem{liu2020self}
Peidong Liu, Joel Janai, Marc Pollefeys, Torsten Sattler, and Andreas Geiger.
\newblock Self-supervised linear motion deblurring.
\newblock {\em IEEE Robotics and Automation Letters}, 5(2):2475--2482, 2020.

\bibitem{nah2019ntire}
Seungjun Nah, Sungyong Baik, Seokil Hong, Gyeongsik Moon, Sanghyun Son, Radu
  Timofte, and Kyoung Mu~Lee.
\newblock Ntire 2019 challenge on video deblurring and super-resolution:
  Dataset and study.
\newblock In {\em CVPRW}, pages 1974--1984, 2019.

\bibitem{niklaus2017video}
Simon Niklaus, Long Mai, and Feng Liu.
\newblock Video frame interpolation via adaptive separable convolution.
\newblock In {\em ICCV}, pages 261--270, 2017.

\bibitem{nimisha2017blur}
Thekke~Madam Nimisha, Akash Kumar~Singh, and Ambasamudram~N Rajagopalan.
\newblock Blur-invariant deep learning for blind-deblurring.
\newblock In {\em ICCV}, pages 4752--4760, 2017.

\bibitem{pan2016blind}
Jinshan Pan, Deqing Sun, Hanspeter Pfister, and Ming-Hsuan Yang.
\newblock Blind image deblurring using dark channel prior.
\newblock In {\em CVPR}, pages 1628--1636, 2016.

\bibitem{pan2020high}
Liyuan Pan, Richard Hartley, Cedric Scheerlinck, Miaomiao Liu, Xin Yu, and
  Yuchao Dai.
\newblock High frame rate video reconstruction based on an event camera.
\newblock {\em IEEE Transactions on Pattern Analysis and Machine Intelligence},
  2020.

\bibitem{pan2020single}
Liyuan Pan, Miaomiao Liu, and Richard Hartley.
\newblock Single image optical flow estimation with an event camera.
\newblock In {\em CVPR}, pages 1669--1678, 2020.

\bibitem{pan2019bringing}
Liyuan Pan, Cedric Scheerlinck, Xin Yu, Richard Hartley, Miaomiao Liu, and
  Yuchao Dai.
\newblock Bringing a blurry frame alive at high frame-rate with an event
  camera.
\newblock In {\em CVPR}, pages 6820--6829, 2019.

\bibitem{purohit2019bringing}
Kuldeep Purohit, Anshul Shah, and AN Rajagopalan.
\newblock Bringing alive blurred moments.
\newblock In {\em CVPR}, pages 6830--6839, 2019.

\bibitem{rebecq2018esim}
Henri Rebecq, Daniel Gehrig, and Davide Scaramuzza.
\newblock Esim: an open event camera simulator.
\newblock In {\em Conference on Robot Learning}, pages 969--982, 2018.

\bibitem{rengarajan2020photosequencing}
Vijay Rengarajan, Shuo Zhao, Ruiwen Zhen, John Glotzbach, Hamid Sheikh, and
  Aswin~C Sankaranarayanan.
\newblock Photosequencing of motion blur using short and long exposures.
\newblock In {\em CVPRW}, pages 510--511, 2020.

\bibitem{scheerlinck2018continuous}
Cedric Scheerlinck, Nick Barnes, and Robert Mahony.
\newblock Continuous-time intensity estimation using event cameras.
\newblock In {\em ACCV}, pages 308--324, 2018.

\bibitem{stoffregen2020reducing}
Timo Stoffregen, Cedric Scheerlinck, Davide Scaramuzza, Tom Drummond, Nick
  Barnes, Lindsay Kleeman, and Robert Mahony.
\newblock Reducing the sim-to-real gap for event cameras.
\newblock In {\em ECCV}, 2020.

\bibitem{sun2013edge}
Libin Sun, Sunghyun Cho, Jue Wang, and James Hays.
\newblock Edge-based blur kernel estimation using patch priors.
\newblock In {\em IEEE International Conference on Computational Photography
  (ICCP)}, pages 1--8, 2013.

\bibitem{wang2020event}
Bishan Wang, Jingwei He, Lei Yu, Gui-Song Xia, and Wen Yang.
\newblock Event enhanced high-quality image recovery.
\newblock In {\em ECCV}, 2020.

\bibitem{wang2020joint}
Zihao~W Wang, Peiqi Duan, Oliver Cossairt, Aggelos Katsaggelos, Tiejun Huang,
  and Boxin Shi.
\newblock Joint filtering of intensity images and neuromorphic events for
  high-resolution noise-robust imaging.
\newblock In {\em CVPR}, pages 1609--1619, 2020.

\bibitem{xu2013unnatural}
Li Xu, Shicheng Zheng, and Jiaya Jia.
\newblock Unnatural $\ell_0$ sparse representation for natural image
  deblurring.
\newblock In {\em CVPR}, pages 1107--1114, 2013.

\bibitem{zhang2015intra}
Haichao Zhang and Jianchao Yang.
\newblock Intra-frame deblurring by leveraging inter-frame camera motion.
\newblock In {\em CVPR}, pages 4036--4044, 2015.

\bibitem{zhou2019spatio}
Shangchen Zhou, Jiawei Zhang, Jinshan Pan, Haozhe Xie, Wangmeng Zuo, and Jimmy
  Ren.
\newblock Spatio-temporal filter adaptive network for video deblurring.
\newblock In {\em ICCV}, pages 2482--2491, 2019.

\bibitem{zhu2018ev}
Alex~Zihao Zhu, Liangzhe Yuan, Kenneth Chaney, and Kostas Daniilidis.
\newblock Ev-flownet: Self-supervised optical flow estimation for event-based
  cameras.
\newblock In {\em Robotics: Science and Systems}, 2018.

\end{thebibliography}
}

\end{document}